\documentclass{article}

\usepackage[margin=1in]{geometry}

\usepackage{microtype}
\usepackage{graphicx}
\usepackage{subcaption}
\usepackage{booktabs}
\usepackage{enumitem}
\usepackage[round,semicolon]{natbib}
\usepackage{xcolor}
\usepackage{textcomp}

\usepackage{hyperref}
\definecolor{darkblue}{rgb}{0, 0.2, 0.7}
\hypersetup{
    colorlinks = true,
    linkcolor = darkblue,
    anchorcolor = darkblue,
    citecolor = darkblue,
    filecolor = darkblue,
    urlcolor = darkblue
}

\usepackage{cleveref}
\usepackage{multicol}

\title{\vspace{-2em}%
  \hrule height 4pt%
  \vskip 0.25in%
  \vskip -\parskip%
  \textbf{
  WT5?! Training Text-to-Text Models to Explain their Predictions
  }%
  \vskip 0.2in%
  \vskip -\parskip%
  \hrule height 1pt%
  \vskip 0.09in}

\author{\textbf{Sharan Narang}\thanks{Equal Contribution. Correspondence to \newline sharannarang@google.com} \qquad  \textbf{Colin Raffel}\footnotemark[1] \qquad \textbf{Katherine Lee} \\ \qquad \textbf{Adam Roberts} \qquad \textbf{Noah Fiedel}  \qquad \textbf{Karishma Malkan} \\ \\ Google Research }

\date{}

\begin{document}
\twocolumn
\maketitle

\begin{abstract}
  Neural networks have recently achieved human-level performance on various challenging natural language processing (NLP) tasks, but it is notoriously difficult to understand \textit{why} a neural network produced a particular prediction.
  In this paper, we leverage the text-to-text framework proposed by \citet{raffel2019exploring} to train language models to output a natural text explanation alongside their prediction.
  Crucially, this requires no modifications to the loss function or training and decoding procedures -- we simply train the model to output the explanation after generating the (natural text) prediction.
  We show that this approach not only obtains state-of-the-art results on ``explainability'' benchmarks, but also permits learning from a limited set of labeled explanations and transferring rationalization abilities across datasets. To facilitate reproducibility and future work, we release our code use to train the models.\footnote{\url{https://github.com/google-research/google-research/tree/master/wt5}}
\end{abstract}

\section{Introduction}

Neural networks excel in a wide variety of practical settings, from computer vision to speech recognition to natural language processing (NLP) and beyond.
In particular, over the past few years it has been shown that large language models pre-trained on an unlabeled text corpus can be subsequently fine-tuned to achieve superhuman performance on NLP tasks that had previously been considered difficult for machines \citep{devlin2018bert,peters2018deep,howard2018universal,lan2019albert,raffel2019exploring}.
It has further recently been shown that all NLP tasks of interest can be cast as a ``text-to-text'' problem \citep{raffel2019exploring}, where the model is fed some text as input and is trained to produce target text as output.
For example, sentiment analysis of a movie review might involve analyzing the input text ``I went to see this movie with my husband, and we both thought the acting was terrible!'' and producing the word ``negative'' to denote a negative sentiment.
This simple and (arguably) universal framework was shown to obtain state-of-the-art results across a variety of NLP tasks.

\begin{figure}[t]
  \begin{center}
    \centerline{\includegraphics[width=0.8\columnwidth]{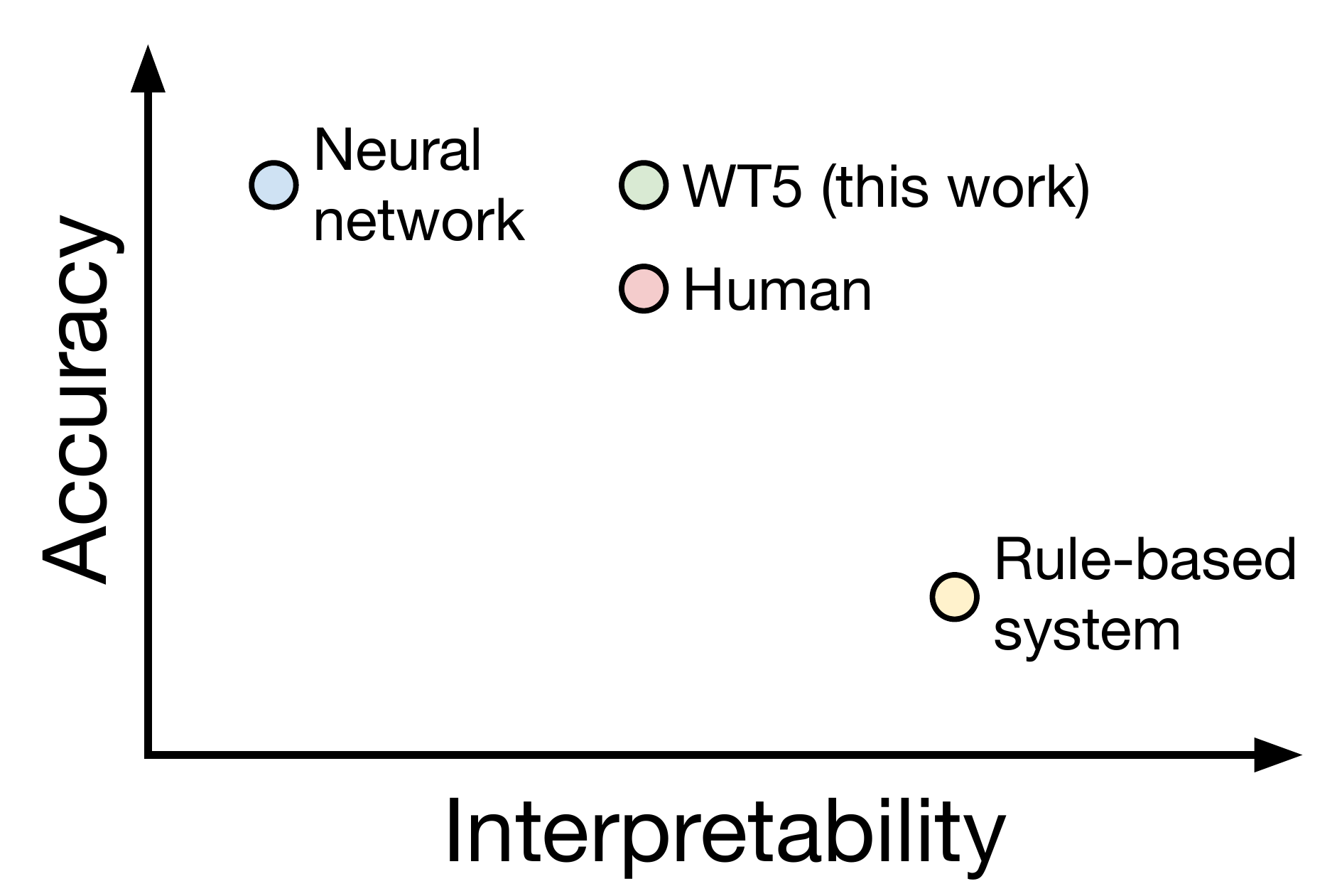}}
    \caption{Illustration of our perspective on the accuracy and interpretability of different models.
    Neural networks (blue) can attain superhuman performance, but are notoriously hard to interpret.
    A rule-based system (yellow) is easy to interpret but rarely performs well on difficult tasks.
    Humans (red) are reasonably accurate and provide some degree of interpretability by being able to verbally explain their predictions.
    In this work, our model (green) is trained both to be highly accurate (in some cases, more accurate than a human) and provide explanations for its predictions as humans do.}
    \label{fig:accuracy_interpretability}
  \end{center}
  \vskip -0.2in
\end{figure}

\begin{figure*}[t]
  \begin{center}
    \centerline{\includegraphics[width=0.8\textwidth]{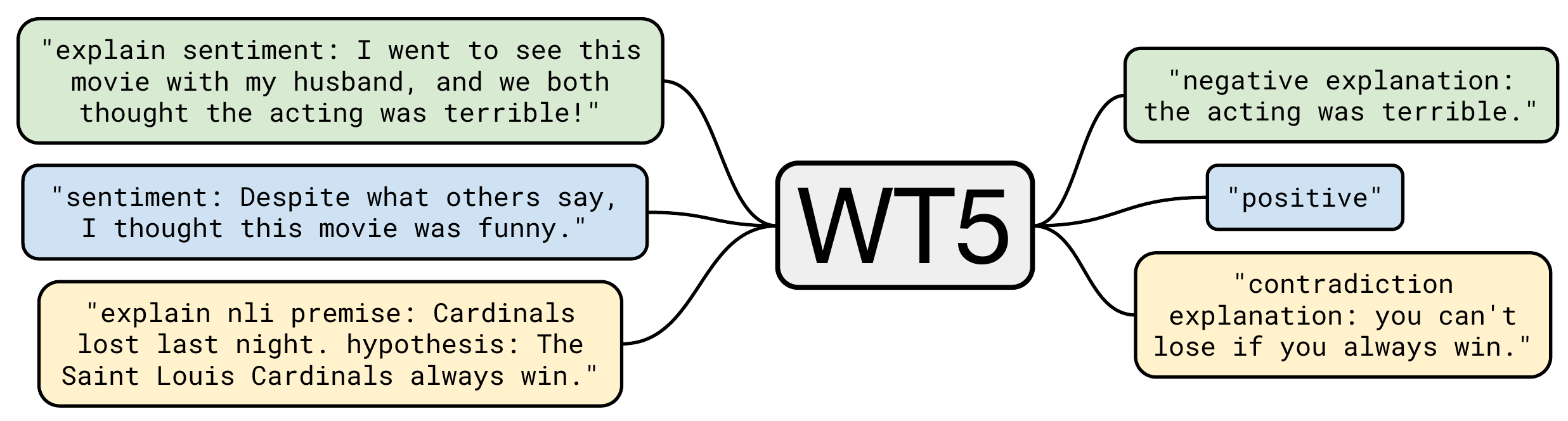}}
    \caption{Diagram of our method for training a text-to-text model to explain its predictions.
    We train the model to generate an explanation when the text ``explain'' is prepended to the input.
    The model can still be trained for classification (without an explanation) simply by omitting the ``explain'' keyword.
    This approach is readily applicable to sentiment analysis, natural language inference (NLI), and other text tasks.
    }
    \label{fig:wt5}
  \end{center}
  \vskip -0.2in
\end{figure*}

In spite of their empirical and practical successes, it is notoriously difficult to determine \textit{why} a neural network has produced a given prediction.
This has led to a substantial body of research that endeavors to make neural networks more ``interpretable'', e.g.\ by attributing its prediction to a given part of its input \citep{baehrens2010explain,sundararajan2017axiomatic,smilkov2017smoothgrad} or by designing architectures that are easier to analyze \citep{foerster2017input,jacobsen2018revnet,bahdanau2014neural,raffel2017online}.
However, the reliability of some of these methods has been questioned \citep{kindermans2019reliability,hooker2018evaluating,jain2019attention,serrano2019attention,pruthi2019learning}, limiting their practical utility.
To motivate the perspective we take in this work, note that we humans (who tend to be rather good at NLP tasks) are also ``black boxes'' in the sense that we cannot obtain a full and transparent view of our own decision-making process.
Instead, we rely on humans to \textit{explain} their judgements.
In contrast, consider a system comprising a list of hard-coded rules (for example, ``if the review contains the word `terrible', the review is negative'').
In this case, it would be simple to understand the behavior and decision-making process of the system, but it's unlikely that such a system would be very accurate (for example, the review ``This movie was anything but terrible!'' suggests a positive sentiment).

Given that humans and neural networks are both highly accurate and hard to interpret, we argue that neural networks should also be given the ability to explain their predictions using natural text.
This idea has already motivated the development of various ``explainability'' datasets in the NLP literature \citep{zaidan2008modeling,camburu2018snli,rajani2019explain,deyoung2019eraser}.
Our main contribution is to show that the text-to-text framework makes it straightforward to train a model to produce an explanation.
Specifically, instead of training the model to simply predict a label (e.g.\ ``negative''), we train it to predict a label and explanation (e.g.\ ``negative explanation: the acting was terrible'').

In addition to producing state-of-the-art results on explainability datasets, this approach also allows for both ``semi-supervised'' training (where explanations are only provided on a subset of the dataset) and for various forms of cross-domain transfer.
For example, we train a model to generate explanations for data from one domain and show that it can generate plausible explanations for out-of-domain data.
From a broad view, we argue that a text-to-text model has an inherent ability to ``communicate'' given its input and output format, and our work mainly involves training these models to communicate better.
We provide a pictorial summary of our perspective on model interpretability in \cref{fig:accuracy_interpretability}.

In the following section, we review the text-to-text framework and corresponding pre-trained model we use and describe our approach in more detail.
\Cref{sec:related} provides an overview of related work on interpretability, particularly for NLP models.
Then, in \cref{sec:experiments}, we introduce the various datasets and evaluation procedures we use for benchmarking before presenting experimental results.
Finally, we conclude with an outlook on the connection between interpretability and training models to communicate with natural language.

\section{Approach}

Before presenting our basic methods, we first review the text-to-text framework we use.
This framework underlies the pre-trained model that we used, which is called the ``Text-to-Text Transfer Transformer'' (T5).
Then, we describe the details of how we fine-tune T5 to produce explanations for its predictions.
We call the resulting model (and our general approach) ``WT5'' as shorthand for ``Why, T5?''.

\subsection{Text-to-Text Framework}

A text-to-text model follows the sequence-to-sequence framework \citep{sutskever2014sequence,kalchbrenner2014convolutional} -- it is fed a sequence of discrete tokens as input and produces a new sequence of tokens as output.
Specifically, the model is provided with an input sequence $\{x_1, \ldots, x_T\}$ and is trained to produce an output sequence $\{y_1, \ldots, y_U\}$ by maximizing $p(y_i | x_1, \ldots, x_T, y_1, \ldots, y_{i - 2}, y_{i - 1} \}$.
At test time, the tokens are sampled from the model's predicted output distribution ($y_i \sim p(y_i | \ldots)$) one at a time and fed back into the model autoregressively until a special end-of-sequence token is generated, at which point the model has produced its complete prediction.
For text problems, the individual sequence elements $y_i$ or $x_i$ are often characters, words, or (in our case), subword token IDs \citep{sennrich2015neural} produced by a tokenizer like SentencePiece \citep{kudo2018subword,kudo2018sentencepiece}.
Notably, \citet{raffel2019exploring} propose converting \textit{all} text problems to the sequence-to-sequence format.
For example, in a classification task, instead of training a classification layer to assign a high probability to some class ID, the model is trained to produce the actual text corresponding to the class label.
Concretely, to train the model to perform sentiment analysis on our running movie review example, the model would be fed the sequence ``sentiment: I went to see this movie with my husband, and we both thought the acting was terrible!'' and would be trained to produce the literal text  ``negative''.
The ``sentiment:'' prefix tells the model what task it should perform, which is useful in multi-task models \citep{caruana1997multitask,ruder2017overview}.

In \citet{raffel2019exploring}, this framework was used to pre-train Transformer \citep{vaswani2017attention} models on a large collection of unlabeled text drawn from the Common Crawl web scrape.
We use the resulting pre-trained models (referred to as T5 for ``Text-to-Text Transfer Transformer'') in our experiments.
Pre-trained models of various sizes are available; we experiment with the ``Base'' model with about $220$ million parameters and the ``11B'' model with around $11$ billion parameters.
Further details on these models and the pre-training procedure are available in \citep{raffel2019exploring}.

\subsection{Generating Explanations}

The text-to-text framework provides a straightforward means of training models to output an explanation alongside their prediction.
We experimented with various ways of modifying the input and output text to include an explanation, and settled on the following recipe:
When we want the model to generate an explanation, we simply prepend the word ``explain'' to the input text and then append ``explanation:'' followed by the explanation to the target text.
In our running movie review example, this produces the input ``explain sentiment: I went to see this movie with my husband, and we both thought the acting was terrible!'' with target ``negative explanation: the acting was terrible.''
Crucially, the model can be simultaneously trained on examples with explanations (which have ``explain'' prepended to their input and ``explanation: ...'' appended to their output) as well as examples with only labels (by omitting ``explain'' from the input and ``explanation: ...'' from the target so that the desired output is only the label text).
This allows us to explore a ``semi-supervised'' setting where we have a dataset that is fully labeled but only a limited number of examples have explanations.
A diagram of this basic approach, with examples for sentiment analysis and natural language inference (NLI) \citep{dagan2005pascal,bowman2015large}, is shown in \cref{fig:wt5}.

\subsection{Extractive Explanations}
\label{sec:extractive}

So far, we have assumed that explanations will be arbitrary text generated by our model.
An alternative way of producing explanations is to train a model to identify spans of text in the input which support its prediction.
This ``extractive'' version is the setting considered by the recent ERASER benchmark \citep{deyoung2019eraser}, which combines various datasets that have been annotated with extractive explanations.
The use of spans makes it possible to use non-generative span-prediction models like BERT \citep{devlin2018bert}.
It also makes evaluation potentially simpler by computing the overlap between the predicted and annotated spans.
In our running movie review example, the explanation text ``the acting was terrible'' appears as a span of text in the input, so this particular example is compatible with the extractive approach.

Note that forcing explanations to be extracted spans is strictly less general.
Consider the task of producing explanations for the Winograd Schema Challenge (WSC) \citep{levesque2012winograd}, where the goal is to disambiguate an ambiguous pronoun.
For example, in the text ``the city government denied the protesters a permit because they feared violence'' the pronoun ``they'' refers to ``the city government'' because governments sometimes fear violence from protesters and not vice-versa.
This explanation for why ``they'' refers to ``the city government'' does not appear anywhere in the text, suggesting that this task (and likely many others) is largely incompatible with extractive explanations.

We include some extractive explanation datasets in our experiments mainly to demonstrate the flexibility of our approach.
To train our model to generate extractive explanations, we include the spans of the input which have been annotated as an explanation with the text ``explanation:'' in the targets and train the model to generate them sequentially.
Then, when the model outputs a prediction and corresponding sequence of explanations, we match each predicted explanation to a span of text in the input, thereby allowing straightforward evaluation using span overlap-based metrics.
A potential issue arises if our model generates an explanation which does not appear in the input text.
We ignore such spurious explanations, though we found this rarely happened in practice.

\section{Related Work}
\label{sec:related}

Measuring and improving the interpretability of neural networks is a heavily-studied area of research; a comprehensive overview is outside of the scope of this work.
Instead, we refer the interested reader to the surveys provided by \citep{doshi2017towards,molnar2019interpretable,guidotti2018survey}.
Most work on interpretability methods focuses on models for computer vision applications (e.g.\ \citet{xu2015show, zhang2018visual}), whereas the interpretability of NLP models is apparently less studied.
A notable exception is the fact that attention-based neural networks \citep{bahdanau2014neural} provide some means of interpretability ``for free'' by examining the weight assigned by the neural network to different regions in the input \citep{graves2013generating,raffel2017online,huang2018music}, but this introspection method has been shown to be unreliable \citep{jain2019attention,serrano2019attention,pruthi2019learning}.
There has separately been significant work on better understanding the behavior NLP models, for example by crafting inputs that cause a misclassification \citep{jia2017adversarial,nie2019adversarial} or diagnosing why they sometimes generate nonsensical text \citep{lee2018hallucinations,belinkov2017synthetic}.

An early investigation into explanations for NLP datasets was performed by \citet{zaidan2008modeling}, who introduced the idea of annotating spans of the input which support the label.
This produced the ``Movie Reviews'' dataset that we consider in our experiments.
The general approach of extractive explanation was recently surveyed and advocated for by \citet{deyoung2019eraser}, who proposed the ERASER benchmark comprising various datasets.
As discussed in \cref{sec:extractive}, our approach is strictly more general in that it also allows us to consider generating abstractive explanations.

\citet{camburu2018snli} have the most philosophically similar perspective to ours.
They consider the generation of abstractive explanations by creating the e-SNLI dataset, which we consider in our experiments. e-SNLI is a variant of the Stanford Natural Language Inference (SNLI) dataset \citep{bowman2015large} that adds human-annotated explanations for all examples in the training, validation, and test sets.
To generate explanations, \citet{camburu2018snli} propose model architectures that generally consist of separate components for classification and explanation.
They consider various tasks, including conditioning the prediction on the explanation and vice versa, as well as producing sentence embeddings.
Most related to this work, they also consider the task of learning to explain with e-SNLI but generating explanations for out-of-domain examples from other natural language inference tasks.
The primary differences between \citet{camburu2018snli} and this work are that our approach requires no special model architecture and that we take advantage of a pre-trained model that is already quite capable of generating natural text.
These differences not only make our method simpler but also produce substantially better performance on the e-SNLI task.

\citet{rajani2019explain} also consider abstractive explanations.
They introduce the CoS-E dataset, which comprises examples from the Commonsense Question Answering (CQA) dataset that have been annotated with abstractive explanations.
However, their focus is mainly on using explanations to improve a model's predictions, and as such they propose first training a model to generate explanations and then training a classifier to produce a prediction based on the original example and the generated explanation.
Interestingly, this provided a substantial performance boost on the CQA dataset.
They include minimal analysis or evaluation of the generated explanations, though they do show (through a few non-cherrypicked examples) that their model can generate explanations for datasets it was not trained on.
The primary focus of our work is on generating useful explanations, so we do not experiment with feeding explanations into a model to improve its predictions.

\section{Experiments}
\label{sec:experiments}

Having introduced our straightforward approach for generating explanations alongside predictions using the text-to-text framework, we now evaluate this idea on the various benchmark datasets described in the following subsection.
In our experiments, we will frequently use human judgements for evaluation because free-form text is notoriously difficult to automatically evaluate and some of the datasets we consider do not have ground-truth explanations.
We describe both the automatic metrics for evaluation used as well as our framework for obtaining human judgements in \cref{sec:evaluation}.
The remainder of this section is devoted to our experimental results.

\subsection{Datasets}
\label{sec:Datasets}

In our experiments, we evaluate on the following datasets:
\textbf{e-SNLI} was proposed by \citet{camburu2018snli}, who annotated every example in the Stanford Natural Language Inference (SNLI) dataset with free-form explanations of the labels.
The natural language inference task involves determining whether a premise entails (implies), contradicts, or has no relationship to a hypothesis.
\textbf{CoS-E} was introduced in \citep{rajani2019explain} and augments the Commonsense Question-Answering (CQA) with free-form explanations.
The CQA task involves answering multiple-choice questions that ostensibly rely on commonsense reasoning or ``world knowledge''.
Note that CoS-E also includes extractive explanations, which we do not use in this paper.
\textbf{Movie Reviews} \citep{zaidan2008modeling} is a sentiment analysis dataset where the goal is to predict whether a movie review has a positive or negative sentiment.
Each review is annotated with spans that support the positive/negative label.
\textbf{MultiRC} \citep{khashabi2018looking} is a reading comprehension dataset that similarly includes spans of the input document supporting the answer to a given question.
Specifically, in MultiRC a model is fed not only a question about a given document but also a candidate answer that the model must then classify as correct or incorrect.
We use the variants of Movie Reviews and MultiRC distributed with the ERASER benchmark \citep{deyoung2019eraser}.

\subsection{Evaluation}
\label{sec:evaluation}

\subsubsection{Quantitative}

All of the datasets we use involve producing a class label based on some input text -- entailment, neutral, or contradiction for e-SNLI, the correct answer from a list of choices for CoS-E, positive or negative for Movie Reviews, and True or False for MultiRC.
As such, for all datasets we report the classification accuracy of our model in order to evaluate the quality of its predictions.

For abstractive explanations, \citet{camburu2018snli} propose using the BLEU score \citep{papineni2002bleu} to compare a predicted explanation against the ground-truth explanation from e-SNLI.
Since \citet{rajani2019explain} mainly consider the setting where explanations are fed as input into a classification model, they do not propose any particular metric for evaluating the quality of generated explanations on CoS-E.
As such, we use the BLEU score both for e-SNLI and CoS-E.
We use SacreBLEU v1.3.0 \citep{post2018call} with “exp” smoothing and “intl” tokenization.
Notably, many of the ground-truth explanations for CoS-E are low quality and/or nonsensical (for example, the question ``Little sarah didn't think that anyone should be kissing boys. She thought that boys had what?'' with answer ``cooties'' was annotated with the explanation ``american horror comedy film directed''; or the question ``What do you fill with ink to print?'' with answer ``printer'' was annotated with the explanation ``health complications'', etc.), suggesting that BLEU scores on CoS-E should be taken with a grain of salt.
We discuss this issue further in \cref{sec:standard}

The ERASER benchmark \citep{deyoung2019eraser} suggests various metrics for measuring whether extracted explanation spans match the ground-truth.
The simplest and most general computes an F1 score based on which entries of the input are labeled as an explanation by the prediction and ground-truth.
Specifically, \citet{deyoung2019eraser} first tokenize the input text using the \texttt{spacy.io} tokenizer,\footnote{\url{http://spacy.io}} and then compute true/false positives/negatives on a token-by-token basis.
We use spacy's ``en\_core\_web\_sm'' model for tokenization to compute the F1 score.

\subsubsection{Qualitative}

The BLEU and F1 scores mentioned above can only loosely measure the quality of an explanation.
We argue that the most reliable way of determining whether an explanation supports a prediction is using human judgement.
The number and scale of our experiments necessitates the use of crowd computing, so we use the Mechanical Turk (MTurk) platform to obtain ratings of model explanations.

Since many of our raters are not language experts, we devised a simple and straightforward set of questions for evaluating a given explanation.
Specifically, we present a rater with the input, predicted label, and explanation and ask whether the explanation supports the label.
We apply this procedure to both abstractive (e-SNLI, CoS-E) and extractive (Movie Reviews, MultiRC) explanations.
For extractive explanations, we present a single answer span at a time alongside the input and predicted label.
Note that this will only allow us to evaluate whether a span is a true or false positive and does not provide a way of evaluating false negatives, but nevertheless provides a helpful perspective on the model's explanation quality.
We provide screenshots of our evaluation forms in \cref{sec:human_study_forms}.

For every dataset we study, we evaluate 100 examples using MTurk with 5 independent ratings for each example.
To ensure quality ratings, we split the 100 examples into batches of 10 and include an attention check (question for which we know the answer) in each group.
If the attention check was failed or not every question was answered, we remove that batch of 10 examples from our analysis and re-rate the batch so that all examples are rated by 5 different raters.
We treat an explanation as correct if the majority of the 5 raters label it as a good explanation.


\subsection{Training Details}

We use the ``Base'' and ``11B'' configurations of T5 in our experiments.
For fine-tuning, we follow the same procedure used for the downstream tasks in \citep{raffel2019exploring}:
As a brief summary, models are fine-tuned trained using the AdaFactor optimizer \citep{shazeer2018adafactor} with a constant learning rate of $0.001$.
We use a batch size of $196{,}608$ tokens for the Base model and $65{,}536$ for 11B.
We use a maximum input sequence length of $512$ for e-SNLI, $128$ for CoS-E, $1024$ for MultiRC and $2048$ for Movie Reviews.
We apply dropout with a rate of $10\%$ throughout fine-tuning.
To obtain model predictions, we perform ``greedy'' decoding by choosing the token ID corresponding to the largest output logit.
For each task, we fine-tune until overfitting is observed on a held-out validation set and choose the checkpoint corresponding to the highest accuracy on the validation set.

\subsection{Results on standard benchmarks}
\label{sec:standard}


\begin{table*}[ht]
    \centering
    \footnotesize
    \caption{Results attained by WT5 and various baselines on the datasets we study.
    Acc is short for accuracy, HE for Human Evaluation, and TF1 for Token F1. F1a is the F1 score over all answers used in the MultiRC dataset \citep{KCRUR18}.
    See \cref{sec:standard} for a description of baselines. Note that for the Human row, Acc and TF1 are measured on our hand-labeled examples while HE is measured on the ground-truth explanations from the dataset. We were not able to run human evaluation for past SoTA models since we do not have access to the explanations produced by those models. $^*$As far as we are aware, the only work which reports accuracy on the Movie Reviews dataset is \citep{zaidan2008modeling}; \citep{deyoung2019eraser} reports an F1 score of $97.4$. Since the Movies Rationale dataset is reasonably class-balanced and models are performing near-perfectly, this F1 score is somewhat comparable to the accuracy scores we report.
    Superscripts denote results from past work: $^a$\cite{liu2019multi}, $^b$\cite{camburu2018snli}, $^c$\cite{lan2019albert}, $^d$\cite{zaidan2008modeling}, $^e$\cite{deyoung2019eraser}, $^f$\cite{raffel2019exploring}, $^g$\cite{bowman2015large}.
    }
    \vskip 0.5em
    \label{tab:standard}
    \begin{tabular}{lccccccccccccc}
    \toprule
        & \multicolumn{3}{c}{e-SNLI} & \multicolumn{3}{c}{CoS-E} & \multicolumn{3}{c}{Movie Reviews} & \multicolumn{3}{c}{MultiRC} \\
         \cmidrule(l{3pt}r{3pt}){2-4}  \cmidrule(l{3pt}r{3pt}){5-7} \cmidrule(l{3pt}r{3pt}){8-10} \cmidrule(l{3pt}r{3pt}){11-13}
                      & Acc    & BLEU   & HE     & Acc    & BLEU   & HE     & Acc    & TF1    & HE     & F1a    & TF1    & HE \\
        Previous SoTA &$91.6^a$&$27.6^b$& --     &$\textbf{83.7}^c$& --     & --     &$92.2^{d*}$&$32.2^e$& --  & $\textbf{87.6}^f$ &$45.6^e$& --     \\
        Human         &$89.0^g$&$22.5^b$& $78.0$ & $80.4$ & $0.51$ & $16.0$ & $100.0$& $29.1$ & $99.0$ & $90.5$ & $51.8$ & $51.0$ \\
        WT5-Base      & $90.9$ & $32.4$ & --     & $59.4$ & $4.63$ & --     & $98.0$ & $\textbf{32.7}$ & --     & $77.8$ & $69.9$ & --     \\
        WT5-11B       & $\textbf{92.3}$ & $\textbf{33.7}$ & $90.0$ & $82.7$ & $\textbf{5.17}$ & $30.0$ & $\textbf{99.0}$ & $31.5$ & $94.0$ & $86.6$ & $\textbf{76.9}$ & $50.0$ \\
        \bottomrule
    \end{tabular}
\end{table*}

\begin{table*}[ht]
    \centering
    \footnotesize
    \caption{\textbf{Non cherry-picked} predictions and explanations produced by WT5-11B on the validation set of each dataset.
    For extractive explanation, we boldface the spans chosen by our model.
    We display a truncated review and passage for the examples from Movie Reviews and MultiRC (respectively) for clarity and space reasons.
    }
    \vskip 0.5em
    \label{tab:examples}
    \begin{tabular}{lp{0.82\textwidth}}
    \toprule
    e-SNLI & \texttt{Premise:} A person in a blue shirt and tan shorts getting ready to roll a bowling ball down the alley. \newline \texttt{Hypothesis:} A person is napping on the couch. \newline \texttt{Predicted label:} contradiction \newline \texttt{Explanation:} A person cannot be napping and getting ready to roll a bowling ball at the same time. \\
    \midrule
    CoS-E & \texttt{Question:} What can you use to store a book while traveling? \newline \texttt{Choices:} library of congress, pocket, backpack, suitcase, synagogue \newline \texttt{Predicted answer:} backpack \newline \texttt{Explanation:} books are often found in backpacks \\
    \midrule
    Movie Reviews & \texttt{Review:} sylvester stallone \textbf{has made some crap films in his lifetime , but this has got to be one of the worst .} a totally \textbf{dull story} that thinks it can use various explosions to make it interesting , " the specialist " is about as exciting as an episode of " dragnet , " and about as well acted . even some attempts at film noir mood are \textbf{destroyed by a sappy script , stupid and unlikable characters , and just plain nothingness} ... \newline \texttt{Predicted label:} negative\\
    \midrule
    MultiRC & \texttt{Passage:} \textbf{Imagine you are standing in a farm field in central Illinois .} The land is so flat you can see for miles and miles . \textbf{On a clear day , you might see a grain silo 20 miles away .} You might think to yourself , it sure is flat around here ... \newline \texttt{Query:} In what part of Illinois might you be able to see a grain silo that is 20 miles away ? \newline
    \texttt{Candidate answer:} Northern Illinois \newline \texttt{Predicted label:} False
 \\
    \bottomrule
    \end{tabular}
\end{table*}

We present WT5-Base and WT5-11B's performance on the standard datasets we consider in \cref{tab:standard}.
All reported results are on the test set except for CoS-E -- human-annotated explanations are not available for the CQA test set, so we report validation set scores instead. Additionally, the test set for MultiRC provided in the Eraser benchmark \citep{deyoung2019eraser} is the validation set from SuperGLUE \citep{wang2019super}. Therefore, the results in this paper do not match the ones reported on the SuperGLUE leader board\footnote{\url{https://super.gluebenchmark.com}/}.
To contextualize these results, we also include the following baselines:
\begin{description}[style=unboxed,leftmargin=0cm]
    \item[Previous State-of-the-art (SoTA)] We report the best score previously achieved on each dataset.
    \item[Human] We estimated human performance on each dataset by hand-labeling examples from the validation set and measuring our accuracy and the corresponding explanation quality score (BLEU or Token F1). We also fed ground-truth labels from each dataset into our human evaluation procedure to get an idea of the quality of explanations in each dataset. For e-SNLI, we use the human agreement scores reported in \citep{camburu2018snli} and \citep{bowman2015large}.
\end{description}

In general, we find that WT5-11B attains the highest scores for its explanations on most of the datasets we studied.
On all datasets, WT5-11B's explanation score is better than the score for the examples we hand-labeled.
This likely does not mean that WT5-11B's explanations are ``better'', but rather that it has learned to model some of the spurious characteristics of ground-truth explanations on a given dataset.
This is borne out in the human evaluation of WT5-11B's explanations, which produced similar scores to the ground-truth explanations on all datasets except for e-SNLI where WT5-11B achieved a $12\%$ higher score.
Separately, WT5-11B attained state-of-the-art accuracy on the e-SNLI and Movie Reviews datasets. For CoS-E and MultiRC, WT5-11B is very close to state-of-the-art accuracy to the T5-11B model which doesn't generate explanations.  
To summarize, our results suggest that WT5-11B is at a human or super-human level at both classifying and explaining examples from the datasets we considered.
We provide some example predictions and explanations produced by WT5-11B in \cref{tab:examples}.

\begin{figure*}[t]
  \begin{center}
    \centerline{\includegraphics[width=\textwidth]{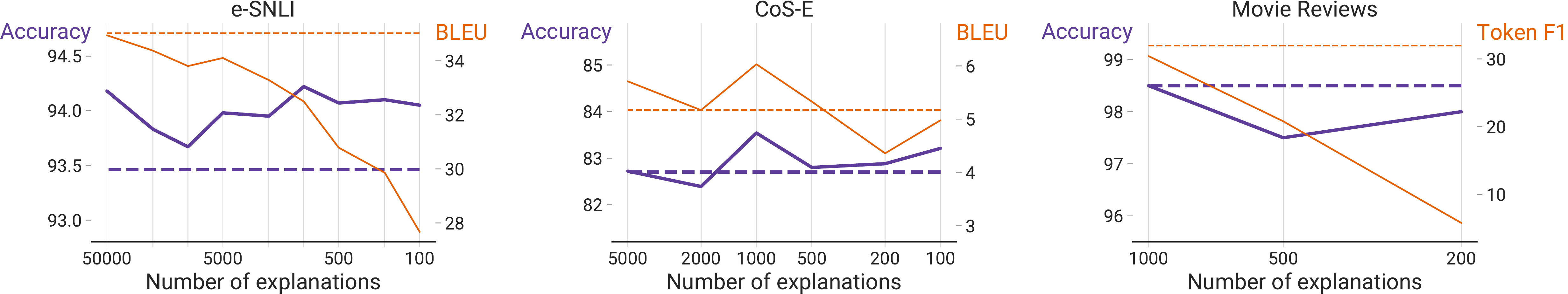}}
    \caption{Accuracy and explanation quality as a function of the number of explanations retained in the training set.
    Dashed lines show the performance attained by using explanations for every example in the training set.
    All scores are reported on the validation set.
    }
    \label{fig:limited}
  \end{center}
  \vskip -0.2in
\end{figure*}

\begin{table*}[ht]
    \centering
    \footnotesize
    \caption{\textbf{Non cherry-picked} predictions and explanations for MNLI, IMDb, and Amazon Reviews based on training WT5-11B. The MNLI explanations are generated using the e-SNLI dataset. The IMDb and Amazon Reviews explanations are generated using Movie Reviews.
    For IMDb and Amazon Reviews, we boldface the explanatory spans chosen by our model. We display only the beginning of the review for IMDb.
    }
    \vskip 0.5em
    \label{tab:transfer_examples}
    \begin{tabular}{lp{0.9\textwidth}}
    \toprule
    MNLI & \texttt{Premise:} He passed these instructions to the pilots that launched at 10:42 and afterward. \newline \texttt{Hypothesis:} Instructions on how to engage were given to the pilot going up. \newline \texttt{Predicted label:} neutral \newline \texttt{Explanation:} The instructions could be about anything, not necessarily how to engage. \\
    \midrule
    IMDb & \texttt{Review:} "Lassie Come Home," "National Velvet," and "The Courage of Lassie," Elizabeth Taylor was eleven years old. Nevertheless, \textbf{her charm and beauty were extraordinary}, and what she lacked in talent and experience was well hidden in \textbf{a fine production that was nominated for five Academy Awards}...
    \newline \texttt{Predicted label:} positive\\
    \midrule
    Amazon & \texttt{Review:} Great TV After having researched the LCD 1080P marketplace extensively, I decided on the Samsung and \textbf{have not been disappointed}. My initial experience, as well as my prior experience with a number of Samsung products makes me confident \textbf{this will prove to be an excellent choice}.
    \newline \texttt{Predicted label:} positive\\
    \bottomrule
    \end{tabular}
\end{table*}

In general, WT5-Base had worse accuracy and explanation quality scores than WT5-11B, but the Base model nevertheless frequently outperformed the previous state-of-the-art and, in some cases, human annotations.
Surprisingly, our hand-annotated explanations achieved a very low BLEU score on CoS-E when evaluated against ground-truth explanations distributed with the dataset.
Upon inspection, we found that this was largely due to the aforementioned noisy and low-quality explanations that are distributed with CoS-E.
This also likely explains why our model's generated explanations were rated correct at a higher rate by MTurk workers than the ground truth explanations provided with CoS-E.


\subsection{Learning from limited explanations}

Our framework facilitates a natural way of learning to generate explanations when only a limited number of examples have been annotated with a ground-truth explanation.
Specifically, if a training example has an annotated explanation, we prepend ``explain'' to the input and append the explanation to the target.
If no explanation has been annotated, we simply train the model to generate the label alone and do not prepend ``explain'' to the input.
These two cases can be seen in the top two examples in \cref{fig:wt5}.
At test time, we ask our model to generate explanations for all of its inputs by prepending ``explain'' to every example.
Hopefully, this approach will produce a model that can generate plausible explanations without requiring much (potentially expensive) hand annotating.

To test this approach, we trained WT5-11B on variants of each dataset after artificially removing most of the annotated explanations (but retaining all labels) and then measured the resulting accuracy and explanation scores on the validation set.
The results for e-SNLI, CoS-E, and Movie Reviews can be seen in \cref{fig:limited}.
For e-SNLI, the accuracy stayed roughly constant but the BLEU score gradually tapered off as the number of explanations decreased.
Nevertheless, with only $100$ explanations, WT5-11B attains a better BLEU score than the previous state-of-the-art \citep{camburu2018snli}.
On CoS-E, the accuracy and BLEU scores both stayed constant (with a mild amount of random fluctuation) even as the number of explanations approached $100$.
We manually inspected the explanations generated by WT5-11B after it had been trained on 100 ground-truth explanations and found that they were indeed of similar quality to those generated by the model trained on all explanations.

In contrast with the results for abstractive explanation, we found that the explanation quality quickly degraded as the number of annotated explanations decreased for both Movie Reviews and MultiRC.
The Movie Reviews dataset comprises only $1{,}600$ training examples; with $1{,}000$ explanations the generated explanation score is reasonable but performance degrades to near-zero as the number of explanations approaches $100$.
On MultiRC, we were able to achieve reasonable results with $10{,}000$ annotated explanations (compared to about $25{,}000$ for the full dataset) but the Token F1 score was $0$ when fewer explanations were used.
This suggests that training the WT5 to generate many extractive explanation spans may require more supervision than training it to generate a single abstractive explanation (as is the case in e-SNLI and CoS-E).

\subsection{Transferring across datasets}

Another way to avoid having to annotate a given dataset with explanations would be to leverage a related dataset for which explanations are already available.
For example, we might use the e-SNLI dataset to train our model to generate explanations and then have the model produce explanations for a different natural language inference dataset.
This also can test whether the model has learned domain-agnostic explanation skills by evaluating performance on a dataset from a different domain.

We evaluated whether WT5-11B could successfully carry out this kind of transfer in two settings. We transferred from e-SNLI to the MNLI dataset \citep{williams2017broad} which measures natural language inference in a much wider variety of domains than SNLI. Secondly, we transferred from Movie Reviews to IMDb \citep{maas2011learning} which consists of a large collection of movie reviews from the website IMDb.
In both cases, we combined all examples from the explanation-annotated and explanation-free datasets and sampled examples randomly from the combined dataset.
For training, we proceeded in a similar fashion as the previous experiment, where we prepended explanation-annotated examples with the word ``explain'' and simply trained the model to predict the label for explanation-free examples.
Since the transfer datasets do not contain ground-truth explanations, our only option for assessing quality was to perform a human study.

In both cases, we found that WT5-11B produced plausible explanations for examples in the dataset which did not have explanations annotated.
Human raters considered $82\%$ of explanations generated from the validation set to be correct for examples from MNLI and $94\%$ for IMDb.
WT5-11B also managed to attain reasonable classification accuracy on each dataset ($91.5\%$ on MNLI and $97.2\%$ on IMDb). We present an example model output for both MNLI and IMDb in \cref{tab:transfer_examples}.

To further test if our model has learnt domain agnostic explanations skills, we evaluated whether WT5-11B could successfully generate explanations for any other kinds of reviews, in addition to movie reviews. To test this transfer, we used the Amazon Reviews dataset \citep{he2016ups,mcauley2015image} which contains reviews for a wide variety of products across diverse categories. The Amazon datasets consists of 5 star ratings as labels. We converted the labels to binary by considering reviews with 1 or 2 stars as positive and those with 4 or 5 stars as negative. The training setup is similar to the one described above for IMDb. The only difference is that we also included examples from Movie Reviews without any explanations. We included Movie Reviews both with and without explanations, so that the model could learn the semantics of the ``explain`` prefix on Movie Reviews and apply it generate explanations for Amazon Reviews. After training, the WT5-11B model produced explanations for almost all examples in the Amazon Reviews dataset, averaging $2.2$ explanations per example. We share some explanations from the Amazon Reviews dataset in table \ref{tab:transfer_examples} and examples from more product categories in \cref{sec:amazon_reviews_explanations}. Additionally, the WT5-11B model achieves a classification accuracy of $98.1\%$ on the Amazon Reviews dataset. This suggests that this form of transfer presents another plausible way of learning to generate explanations on a dataset where they have not yet been annotated, even when the dataset is from another domain.

\subsection{Transferring across tasks}

To push the idea of transferring explanation abilities further, we tested whether WT5-11B could successfully transfer across \textit{tasks}. Specifically, we evaluated whether WT5-11B could be trained to generate explanations on e-SNLI and then produce plausible explanations for the CoS-E dataset.

Note that e-SNLI and CoS-E are quite different tasks:
For e-SNLI, the model receives a premise and hypothesis and must produce an NLI label from three fixed classes; for CoS-E, it is fed a question and must answer  given five example-specific choices.
Given these differences, we modified the task processing to better align the formatting.
In CoS-E, we modified the tokens ``cos\_e'' and ``question:'' to ``nli'' and ``premise:'' respectively.
We also modified the e-SNLI inputs to contain three fixed choices: ``entailment'', ``neutral'', and ``contradiction''.
We ablated these changes and found that both were necessary for our models to learn to generate explanations for CoS-E.
Additionally, we found that decoding with beam-search improved the reliability of generating explanations.

Similar to the procedure for Movie Reviews to Amazon Reviews transfer, we fine-tuned WT5-11B on a mixture of the full e-SNLI dataset with explanations, the full e-SNLI dataset without explanations, and the CoS-E dataset without explanations. We applied this model to generate ``zero-shot'' explanations for CoS-E.
After training, our model produced an explanation for most examples from the CoS-E validation set.
We share some cherry-picked CoS-E explanations generated by our model in \cref{tab:transfer-examples}.
While our model did not 100\% reliably transfer explanation abilities across tasks, this result establishes that zero-shot explanation transfer across tasks is indeed possible and motivates future work on task-agnostic explanation abilities.

\begin{table}[t]
    \footnotesize
    \caption{\textbf{Cherry-picked} predictions and explanations produced by WT5-11B when transferring from e-SNLI to CoS-E. Multiple-choice answers omitted for brevity. The word ``\textit{nice}'' in the third example is a typo that appears in the dataset.
    }
    \vskip 0.5em
    \label{tab:transfer-examples}
    \begin{tabular}{p{0.94\columnwidth}}
    \toprule
    \texttt{Question:} When going to sleep what happens to your body? \\
    \texttt{Predicted answer:} relaxation \\ 
    \texttt{Explanation:} When going to sleep, your body goes into a state of relaxation. \\
    \midrule
    \texttt{Question:} What rubber toy filled with helium will make a child happy? \\
    \texttt{Predicted answer:} balloon \\ 
    \texttt{Explanation:} balloon is a toy filled with helium. \\
    \midrule
    \texttt{Question:} James's nice asked him about her grandfather. She was interested in learning about what? \\
    \texttt{Predicted answer:} family tree \\ 
    \texttt{Explanation:} A family tree is a way of tracing a family's lineage. \\
    \midrule
    \texttt{Question:} Where can one obtain a bass fiddle? \\
    \texttt{Predicted answer:} music store \\ 
    \texttt{Explanation:} A bass fiddle is an instrument. \\
    \bottomrule
    \end{tabular}
\end{table}


\section{Conclusion}

In this paper, we demonstrated how the text-to-text framework can be straightforwardly used to train a model to generate explanations for its predictions.
The resulting model, called WT5-11B, achieved state-of-the-art results on a diverse collection of benchmark datasets and in many cases matched human abilities in both classification performance and explanation abilities.
We also showed how this approach facilitates learning from limited labeled explanations and transferring explanatory capabilities across domains and tasks.

At a high level, our results can be seen as a small step towards improving our models' abilities to communicate.
For example, sentiment analysis on the Movie Reviews dataset is loosely equivalent to asking the model ``what is the sentiment of this movie review?'' and our work allows us to further ask the model ``why?''.
While we are broadly interested in making models communicate more naturally, we also recognize that this approach provides only a surface-level improvement of interpretability:
Much like humans, our approach does not guarantee that the produced explanation actually explains the specific reasons why a model generated its prediction.
In other words, the model could potentially just make up a reasonable-sounding explanation instead of providing a truly accurate description of its causal decision-making process.
Nevertheless, we are excited to see the field progress more towards more human-like text models.

\bibliography{paper}

\begin{thebibliography}{54}
\providecommand{\natexlab}[1]{#1}
\providecommand{\url}[1]{\texttt{#1}}
\expandafter\ifx\csname urlstyle\endcsname\relax
  \providecommand{\doi}[1]{doi: #1}\else
  \providecommand{\doi}{doi: \begingroup \urlstyle{rm}\Url}\fi

\bibitem[Baehrens et~al.(2010)Baehrens, Schroeter, Harmeling, Kawanabe, Hansen,
  and M{\~A}{\v{z}}ller]{baehrens2010explain}
Baehrens, D., Schroeter, T., Harmeling, S., Kawanabe, M., Hansen, K., and
  M{\~A}{\v{z}}ller, K.-R.
\newblock How to explain individual classification decisions.
\newblock \emph{Journal of Machine Learning Research}, 11\penalty0 (June),
  2010.

\bibitem[Bahdanau et~al.(2014)Bahdanau, Cho, and Bengio]{bahdanau2014neural}
Bahdanau, D., Cho, K., and Bengio, Y.
\newblock Neural machine translation by jointly learning to align and
  translate.
\newblock \emph{arXiv preprint arXiv:1409.0473}, 2014.

\bibitem[Belinkov \& Bisk(2017)Belinkov and Bisk]{belinkov2017synthetic}
Belinkov, Y. and Bisk, Y.
\newblock Synthetic and natural noise both break neural machine translation.
\newblock \emph{arXiv preprint arXiv:1711.02173}, 2017.

\bibitem[Bowman et~al.(2015)Bowman, Angeli, Potts, and
  Manning]{bowman2015large}
Bowman, S.~R., Angeli, G., Potts, C., and Manning, C.~D.
\newblock A large annotated corpus for learning natural language inference.
\newblock \emph{arXiv preprint arXiv:1508.05326}, 2015.

\bibitem[Camburu et~al.(2018)Camburu, Rockt{\"a}schel, Lukasiewicz, and
  Blunsom]{camburu2018snli}
Camburu, O.-M., Rockt{\"a}schel, T., Lukasiewicz, T., and Blunsom, P.
\newblock e-snli: Natural language inference with natural language
  explanations.
\newblock In \emph{Advances in Neural Information Processing Systems}, 2018.

\bibitem[Caruana(1997)]{caruana1997multitask}
Caruana, R.
\newblock Multitask learning.
\newblock \emph{Machine learning}, 28\penalty0 (1), 1997.

\bibitem[Dagan et~al.(2005)Dagan, Glickman, and Magnini]{dagan2005pascal}
Dagan, I., Glickman, O., and Magnini, B.
\newblock The pascal recognising textual entailment challenge.
\newblock In \emph{Machine Learning Challenges Workshop}, pp.\  177--190.
  Springer, 2005.

\bibitem[Devlin et~al.(2018)Devlin, Chang, Lee, and Toutanova]{devlin2018bert}
Devlin, J., Chang, M.-W., Lee, K., and Toutanova, K.
\newblock {BERT}: Pre-training of deep bidirectional transformers for language
  understanding.
\newblock \emph{arXiv preprint arXiv:1810.04805}, 2018.

\bibitem[DeYoung et~al.(2019)DeYoung, Jain, Rajani, Lehman, Xiong, Socher, and
  Wallace]{deyoung2019eraser}
DeYoung, J., Jain, S., Rajani, N.~F., Lehman, E., Xiong, C., Socher, R., and
  Wallace, B.~C.
\newblock Eraser: A benchmark to evaluate rationalized nlp models.
\newblock \emph{arXiv preprint arXiv:1911.03429}, 2019.

\bibitem[Doshi-Velez \& Kim(2017)Doshi-Velez and Kim]{doshi2017towards}
Doshi-Velez, F. and Kim, B.
\newblock Towards a rigorous science of interpretable machine learning.
\newblock \emph{arXiv preprint arXiv:1702.08608}, 2017.

\bibitem[Foerster et~al.(2017)Foerster, Gilmer, Sohl-Dickstein, Chorowski, and
  Sussillo]{foerster2017input}
Foerster, J.~N., Gilmer, J., Sohl-Dickstein, J., Chorowski, J., and Sussillo,
  D.
\newblock Input switched affine networks: an rnn architecture designed for
  interpretability.
\newblock In \emph{Proceedings of the 34th International Conference on Machine
  Learning}, 2017.

\bibitem[Graves(2013)]{graves2013generating}
Graves, A.
\newblock Generating sequences with recurrent neural networks.
\newblock \emph{arXiv preprint arXiv:1308.0850}, 2013.

\bibitem[Guidotti et~al.(2018)Guidotti, Monreale, Ruggieri, Turini, Giannotti,
  and Pedreschi]{guidotti2018survey}
Guidotti, R., Monreale, A., Ruggieri, S., Turini, F., Giannotti, F., and
  Pedreschi, D.
\newblock A survey of methods for explaining black box models.
\newblock \emph{ACM computing surveys (CSUR)}, 51\penalty0 (5), 2018.

\bibitem[He \& McAuley(2016)He and McAuley]{he2016ups}
He, R. and McAuley, J.
\newblock Ups and downs: Modeling the visual evolution of fashion trends with
  one-class collaborative filtering.
\newblock In \emph{proceedings of the 25th international conference on world
  wide web}, pp.\  507--517, 2016.

\bibitem[Hooker et~al.(2018)Hooker, Erhan, Kindermans, and
  Kim]{hooker2018evaluating}
Hooker, S., Erhan, D., Kindermans, P.-J., and Kim, B.
\newblock Evaluating feature importance estimates.
\newblock \emph{arXiv preprint arXiv:1806.10758}, 2018.

\bibitem[Howard \& Ruder(2018)Howard and Ruder]{howard2018universal}
Howard, J. and Ruder, S.
\newblock Universal language model fine-tuning for text classification.
\newblock \emph{arXiv preprint arXiv:1801.06146}, 2018.

\bibitem[Huang et~al.(2018)Huang, Vaswani, Uszkoreit, Shazeer, Hawthorne, Dai,
  Hoffman, and Eck]{huang2018music}
Huang, C.-Z.~A., Vaswani, A., Uszkoreit, J., Shazeer, N., Hawthorne, C., Dai,
  A.~M., Hoffman, M.~D., and Eck, D.
\newblock Music transformer: Generating music with long-term structure.
\newblock \emph{arXiv preprint arXiv:1809.04281}, 2018.

\bibitem[Jacobsen et~al.(2018)Jacobsen, Smeulders, and
  Oyallon]{jacobsen2018revnet}
Jacobsen, J.-H., Smeulders, A., and Oyallon, E.
\newblock i-revnet: Deep invertible networks.
\newblock \emph{arXiv preprint arXiv:1802.07088}, 2018.

\bibitem[Jain \& Wallace(2019)Jain and Wallace]{jain2019attention}
Jain, S. and Wallace, B.~C.
\newblock Attention is not explanation.
\newblock \emph{arXiv preprint arXiv:1902.10186}, 2019.

\bibitem[Jia \& Liang(2017)Jia and Liang]{jia2017adversarial}
Jia, R. and Liang, P.
\newblock Adversarial examples for evaluating reading comprehension systems.
\newblock \emph{arXiv preprint arXiv:1707.07328}, 2017.

\bibitem[Kalchbrenner et~al.(2014)Kalchbrenner, Grefenstette, and
  Blunsom]{kalchbrenner2014convolutional}
Kalchbrenner, N., Grefenstette, E., and Blunsom, P.
\newblock A convolutional neural network for modelling sentences.
\newblock \emph{arXiv preprint arXiv:1404.2188}, 2014.

\bibitem[Khashabi et~al.(2018{\natexlab{a}})Khashabi, Chaturvedi, Roth,
  Upadhyay, and Roth]{KCRUR18}
Khashabi, D., Chaturvedi, S., Roth, M., Upadhyay, S., and Roth, D.
\newblock {Looking Beyond the Surface: A Challenge Set for Reading
  Comprehension over Multiple Sentences}.
\newblock In \emph{Proc. of the Annual Conference of the North American Chapter
  of the Association for Computational Linguistics (NAACL)},
  2018{\natexlab{a}}.
\newblock URL \url{http://cogcomp.org/papers/2018-MultiRC-NAACL.pdf}.

\bibitem[Khashabi et~al.(2018{\natexlab{b}})Khashabi, Chaturvedi, Roth,
  Upadhyay, and Roth]{khashabi2018looking}
Khashabi, D., Chaturvedi, S., Roth, M., Upadhyay, S., and Roth, D.
\newblock Looking beyond the surface: A challenge set for reading comprehension
  over multiple sentences.
\newblock In \emph{Proceedings of the 2018 Conference of the North American
  Chapter of the Association for Computational Linguistics},
  2018{\natexlab{b}}.

\bibitem[Kindermans et~al.(2019)Kindermans, Hooker, Adebayo, Alber, Sch{\"u}tt,
  D{\"a}hne, Erhan, and Kim]{kindermans2019reliability}
Kindermans, P.-J., Hooker, S., Adebayo, J., Alber, M., Sch{\"u}tt, K.~T.,
  D{\"a}hne, S., Erhan, D., and Kim, B.
\newblock The (un) reliability of saliency methods.
\newblock In \emph{Explainable AI: Interpreting, Explaining and Visualizing
  Deep Learning}. 2019.

\bibitem[Kudo(2018)]{kudo2018subword}
Kudo, T.
\newblock Subword regularization: Improving neural network translation models
  with multiple subword candidates.
\newblock \emph{arXiv preprint arXiv:1804.10959}, 2018.

\bibitem[Kudo \& Richardson(2018)Kudo and Richardson]{kudo2018sentencepiece}
Kudo, T. and Richardson, J.
\newblock Sentencepiece: A simple and language independent subword tokenizer
  and detokenizer for neural text processing.
\newblock \emph{arXiv preprint arXiv:1808.06226}, 2018.

\bibitem[Lan et~al.(2019)Lan, Chen, Goodman, Gimpel, Sharma, and
  Soricut]{lan2019albert}
Lan, Z., Chen, M., Goodman, S., Gimpel, K., Sharma, P., and Soricut, R.
\newblock {ALBERT}: A lite bert for self-supervised learning of language
  representations.
\newblock \emph{arXiv preprint arXiv:1909.11942}, 2019.

\bibitem[Lee et~al.(2018)Lee, Firat, Agarwal, Fannjiang, and
  Sussillo]{lee2018hallucinations}
Lee, K., Firat, O., Agarwal, A., Fannjiang, C., and Sussillo, D.
\newblock Hallucinations in neural machine translation.
\newblock In \emph{NeurIPS Workshop on Interpretability and Robustness in
  Audio, Speech, and Language}, 2018.

\bibitem[Levesque et~al.(2012)Levesque, Davis, and
  Morgenstern]{levesque2012winograd}
Levesque, H., Davis, E., and Morgenstern, L.
\newblock The winograd schema challenge.
\newblock In \emph{Thirteenth International Conference on the Principles of
  Knowledge Representation and Reasoning}, 2012.

\bibitem[Liu et~al.(2019)Liu, He, Chen, and Gao]{liu2019multi}
Liu, X., He, P., Chen, W., and Gao, J.
\newblock Multi-task deep neural networks for natural language understanding.
\newblock \emph{arXiv preprint arXiv:1901.11504}, 2019.

\bibitem[Maas et~al.(2011)Maas, Daly, Pham, Huang, Ng, and
  Potts]{maas2011learning}
Maas, A.~L., Daly, R.~E., Pham, P.~T., Huang, D., Ng, A.~Y., and Potts, C.
\newblock Learning word vectors for sentiment analysis.
\newblock In \emph{Proceedings of the 49th annual meeting of the association
  for computational linguistics}, 2011.

\bibitem[McAuley et~al.(2015)McAuley, Targett, Shi, and Van
  Den~Hengel]{mcauley2015image}
McAuley, J., Targett, C., Shi, Q., and Van Den~Hengel, A.
\newblock Image-based recommendations on styles and substitutes.
\newblock In \emph{Proceedings of the 38th International ACM SIGIR Conference
  on Research and Development in Information Retrieval}, pp.\  43--52, 2015.

\bibitem[Molnar(2019)]{molnar2019interpretable}
Molnar, C.
\newblock \emph{Interpretable machine learning}.
\newblock 2019.

\bibitem[Nie et~al.(2019)Nie, Williams, Dinan, Bansal, Weston, and
  Kiela]{nie2019adversarial}
Nie, Y., Williams, A., Dinan, E., Bansal, M., Weston, J., and Kiela, D.
\newblock Adversarial nli: A new benchmark for natural language understanding.
\newblock \emph{arXiv preprint arXiv:1910.14599}, 2019.

\bibitem[Papineni et~al.(2002)Papineni, Roukos, Ward, and
  Zhu]{papineni2002bleu}
Papineni, K., Roukos, S., Ward, T., and Zhu, W.-J.
\newblock Bleu: a method for automatic evaluation of machine translation.
\newblock In \emph{Proceedings of the 40th annual meeting on association for
  computational linguistics}, 2002.

\bibitem[Peters et~al.(2018)Peters, Neumann, Iyyer, Gardner, Clark, Lee, and
  Zettlemoyer]{peters2018deep}
Peters, M.~E., Neumann, M., Iyyer, M., Gardner, M., Clark, C., Lee, K., and
  Zettlemoyer, L.
\newblock Deep contextualized word representations.
\newblock \emph{arXiv preprint arXiv:1802.05365}, 2018.

\bibitem[Post(2018)]{post2018call}
Post, M.
\newblock A call for clarity in reporting bleu scores.
\newblock \emph{arXiv preprint arXiv:1804.08771}, 2018.

\bibitem[Pruthi et~al.(2019)Pruthi, Gupta, Dhingra, Neubig, and
  Lipton]{pruthi2019learning}
Pruthi, D., Gupta, M., Dhingra, B., Neubig, G., and Lipton, Z.~C.
\newblock Learning to deceive with attention-based explanations.
\newblock \emph{arXiv preprint arXiv:1909.07913}, 2019.

\bibitem[Raffel et~al.(2017)Raffel, Luong, Liu, Weiss, and
  Eck]{raffel2017online}
Raffel, C., Luong, M.-T., Liu, P.~J., Weiss, R.~J., and Eck, D.
\newblock Online and linear-time attention by enforcing monotonic alignments.
\newblock In \emph{Proceedings of the 34th International Conference on Machine
  Learning}, 2017.

\bibitem[Raffel et~al.(2019)Raffel, Shazeer, Roberts, Lee, Narang, Matena,
  Zhou, Li, and Liu]{raffel2019exploring}
Raffel, C., Shazeer, N., Roberts, A., Lee, K., Narang, S., Matena, M., Zhou,
  Y., Li, W., and Liu, P.~J.
\newblock Exploring the limits of transfer learning with a unified text-to-text
  transformer.
\newblock \emph{arXiv preprint arXiv:1910.10683}, 2019.

\bibitem[Rajani et~al.(2019)Rajani, McCann, Xiong, and
  Socher]{rajani2019explain}
Rajani, N.~F., McCann, B., Xiong, C., and Socher, R.
\newblock Explain yourself! leveraging language models for commonsense
  reasoning.
\newblock \emph{arXiv preprint arXiv:1906.02361}, 2019.

\bibitem[Ruder(2017)]{ruder2017overview}
Ruder, S.
\newblock An overview of multi-task learning in deep neural networks.
\newblock \emph{arXiv preprint arXiv:1706.05098}, 2017.

\bibitem[Sennrich et~al.(2015)Sennrich, Haddow, and Birch]{sennrich2015neural}
Sennrich, R., Haddow, B., and Birch, A.
\newblock Neural machine translation of rare words with subword units.
\newblock \emph{arXiv preprint arXiv:1508.07909}, 2015.

\bibitem[Serrano \& Smith(2019)Serrano and Smith]{serrano2019attention}
Serrano, S. and Smith, N.~A.
\newblock Is attention interpretable?
\newblock \emph{arXiv preprint arXiv:1906.03731}, 2019.

\bibitem[Shazeer \& Stern(2018)Shazeer and Stern]{shazeer2018adafactor}
Shazeer, N. and Stern, M.
\newblock Adafactor: Adaptive learning rates with sublinear memory cost.
\newblock \emph{arXiv preprint arXiv:1804.04235}, 2018.

\bibitem[Smilkov et~al.(2017)Smilkov, Thorat, Kim, Vi{\'e}gas, and
  Wattenberg]{smilkov2017smoothgrad}
Smilkov, D., Thorat, N., Kim, B., Vi{\'e}gas, F., and Wattenberg, M.
\newblock Smoothgrad: removing noise by adding noise.
\newblock \emph{arXiv preprint arXiv:1706.03825}, 2017.

\bibitem[Sundararajan et~al.(2017)Sundararajan, Taly, and
  Yan]{sundararajan2017axiomatic}
Sundararajan, M., Taly, A., and Yan, Q.
\newblock Axiomatic attribution for deep networks.
\newblock In \emph{Proceedings of the 34th International Conference on Machine
  Learning}, 2017.

\bibitem[Sutskever et~al.(2014)Sutskever, Vinyals, and
  Le]{sutskever2014sequence}
Sutskever, I., Vinyals, O., and Le, Q.~V.
\newblock Sequence to sequence learning with neural networks.
\newblock In \emph{Advances in neural information processing systems}, 2014.

\bibitem[Vaswani et~al.(2017)Vaswani, Shazeer, Parmar, Uszkoreit, Jones, Gomez,
  Kaiser, and Polosukhin]{vaswani2017attention}
Vaswani, A., Shazeer, N., Parmar, N., Uszkoreit, J., Jones, L., Gomez, A.~N.,
  Kaiser, {\L}., and Polosukhin, I.
\newblock Attention is all you need.
\newblock In \emph{Advances in neural information processing systems}, 2017.

\bibitem[Wang et~al.(2019)Wang, Pruksachatkun, Nangia, Singh, Michael, Hill,
  Levy, and Bowman]{wang2019super}
Wang, A., Pruksachatkun, Y., Nangia, N., Singh, A., Michael, J., Hill, F.,
  Levy, O., and Bowman, S.
\newblock Superglue: A stickier benchmark for general-purpose language
  understanding systems.
\newblock In Wallach, H., Larochelle, H., Beygelzimer, A., d\textquotesingle
  Alch\'{e}-Buc, F., Fox, E., and Garnett, R. (eds.), \emph{Advances in Neural
  Information Processing Systems 32}, pp.\  3266--3280. Curran Associates,
  Inc., 2019.
\newblock URL
  \url{http://papers.nips.cc/paper/8589-superglue-a-stickier-benchmark-for-general-purpose-language-understanding-systems.pdf}.

\bibitem[Williams et~al.(2017)Williams, Nangia, and Bowman]{williams2017broad}
Williams, A., Nangia, N., and Bowman, S.~R.
\newblock A broad-coverage challenge corpus for sentence understanding through
  inference.
\newblock \emph{arXiv preprint arXiv:1704.05426}, 2017.

\bibitem[Xu et~al.(2015)Xu, Ba, Kiros, Cho, Courville, Salakhudinov, Zemel, and
  Bengio]{xu2015show}
Xu, K., Ba, J., Kiros, R., Cho, K., Courville, A., Salakhudinov, R., Zemel, R.,
  and Bengio, Y.
\newblock Show, attend and tell: Neural image caption generation with visual
  attention.
\newblock In \emph{International conference on machine learning}, 2015.

\bibitem[Zaidan \& Eisner(2008)Zaidan and Eisner]{zaidan2008modeling}
Zaidan, O. and Eisner, J.
\newblock Modeling annotators: A generative approach to learning from annotator
  rationales.
\newblock In \emph{Proceedings of the 2008 conference on Empirical methods in
  natural language processing}, pp.\  31--40, 2008.

\bibitem[Zhang \& Zhu(2018)Zhang and Zhu]{zhang2018visual}
Zhang, Q. and Zhu, S.
\newblock Visual interpretability for deep learning: a survey.
\newblock \emph{Frontiers of Information Technology and Electronic
  Engineering}, 19, 02 2018.
\newblock \doi{10.1631/FITEE.1700808}.

\end{thebibliography}
\bibliographystyle{icml2020}

\clearpage
\newpage
\onecolumn
\appendix
\section{Amazon Reviews explanations}\label{sec:amazon_reviews_explanations}

\begin{table*}[ht]
    \centering
    \footnotesize
    \caption{\textbf{Non cherry-picked} predictions and explanations for Amazon Reviews based on training WT5-11B for different product categories. We boldface the explanatory spans chosen by our model.
    }
    \vskip 0.5em
    \label{tab:amazon_reviews_examples}
    \begin{tabular}{lp{0.8\textwidth}}
    \toprule
    Product Category & Review \\
    \midrule
    Apparel & Lovely vest Fits beautifully (or rather it did before my husband lost 70 pounds), \textbf{true to size}. He wore this a lot, so it went through the washer several times, and \textbf{still looks great}. Very soft material, has not pilled or faded.
    \newline \texttt{Predicted label:} positive\\
    \midrule
    Books & \textbf{a must if you wanna create comics/manga this teaches you everything you need to know}, from paneling, to creating believeable characters, to perspictive, and \textbf{covers everything pretty much}...
    \newline \texttt{Predicted label:} positive\\
    \midrule
    Luggage & pretty good So far I've only used this bag a couple of times but \textbf{it has served it's purpose}. It fits in a standard overhead storage bin, \textbf{I love the bright royal blue color, and it appears to be well-made}. My only complaint is that the extension handle got stuck after only using it a couple of times. Otherwise, \textbf{this is a great piece of luggage}.
    \newline \texttt{Predicted label:} positive\\
    \midrule
    Musical instruments & Worked for about 2 weeks! This product is very poorly made.  My kids, ages 2.5 and 4, got the item for Christmas and were able to enjoy it for approximately 2 weeks before \textbf{the microphone completely stopped working}.  They were not hard on the product at all - I watched them play with it every time.  There is absolutely no reason it should have stopped working.  It is basically now trash.  \textbf{I definitely do not recommend this product} if you want a functioning microphone!!
    \newline \texttt{Predicted label:} negative\\
    \midrule
    Office products & \textbf{Stay away!} I guess you get what you pay for. Basically, I installed the 3 colors, and the small black. The yellow didn't work. I tried cleaning the contacts and the yellow worked sometimes, never well. Then \textbf{the magenta stopped working}. Total junk!
    \newline \texttt{Predicted label:} negative\\
    \midrule
    Outdoors & \textbf{Highly recommended Awesome}  ... \textbf{switch is a little confusing at first} ... Hope they hold up ... have not yet tossed them in the service truck .. \textbf{purchase primarily because of the lifetime warranty}.  light it bright
    \newline \texttt{Predicted label:} positive\\
    \midrule
    Shoes & \textbf{Beware}. Replicas. \textbf{Not genuine}. The gold mirror coating rubbed off while cleaning the lenses -- while using the enclosed cloth. (See photo with spot on lens and paint on cloth.) After doing a bit of research I've come to the conclusion that these are fakes.
    \newline \texttt{Predicted label:} negative\\    
    \midrule
    Toys & Beautiful float, but \textbf{deflates rapidly}... Float looks as advertised; however, it \textbf{takes considerable time to pump up, and then deflates within a few days... Sigh}*
    \newline \texttt{Predicted label:} negative\\
    \bottomrule
    \end{tabular}
\end{table*}
\newpage
\section{Human Study GUIs}\label{sec:human_study_forms}

\begin{figure*}[ht]
  \begin{center}
    \centerline{\includegraphics[width=0.8\textwidth]{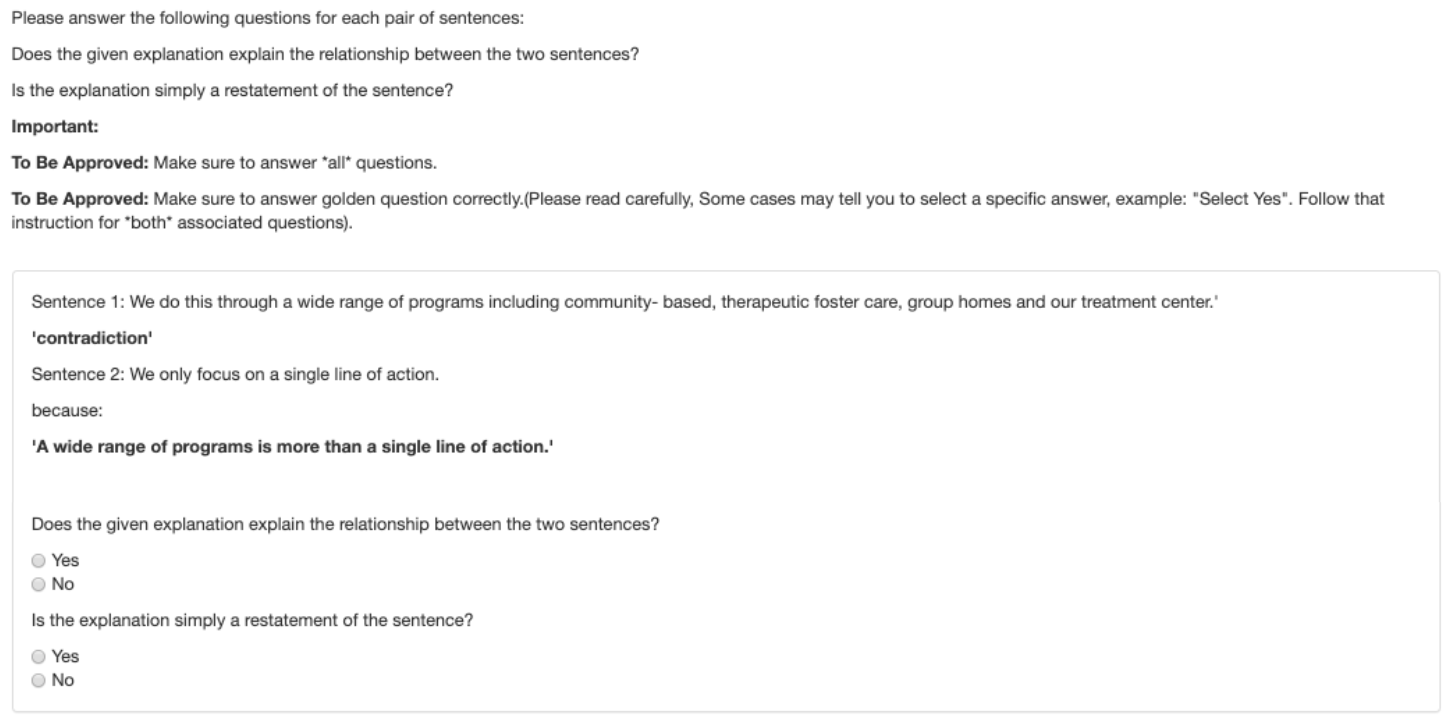}}
    \caption{GUI for MNLI (and also e-SNLI). The explanation provided is a generated by WT5-11B.}
    \label{fig:gui-mnli}
  \end{center}
  \vskip -0.2in
\end{figure*}

\begin{figure*}[ht]
  \begin{center}
    \includegraphics[width=0.8\textwidth]{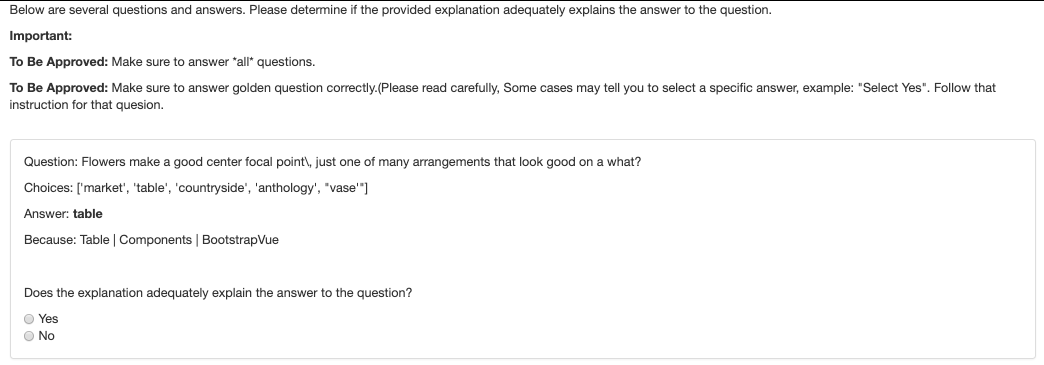}
    \caption{GUI for CoS-E. The explanation is from the validation set of the dataset.}
    \label{fig:gui-cose}
  \end{center}
  \vskip -0.2in
\end{figure*}

\begin{figure*}[ht]
  \begin{center}
    \includegraphics[width=0.8\textwidth]{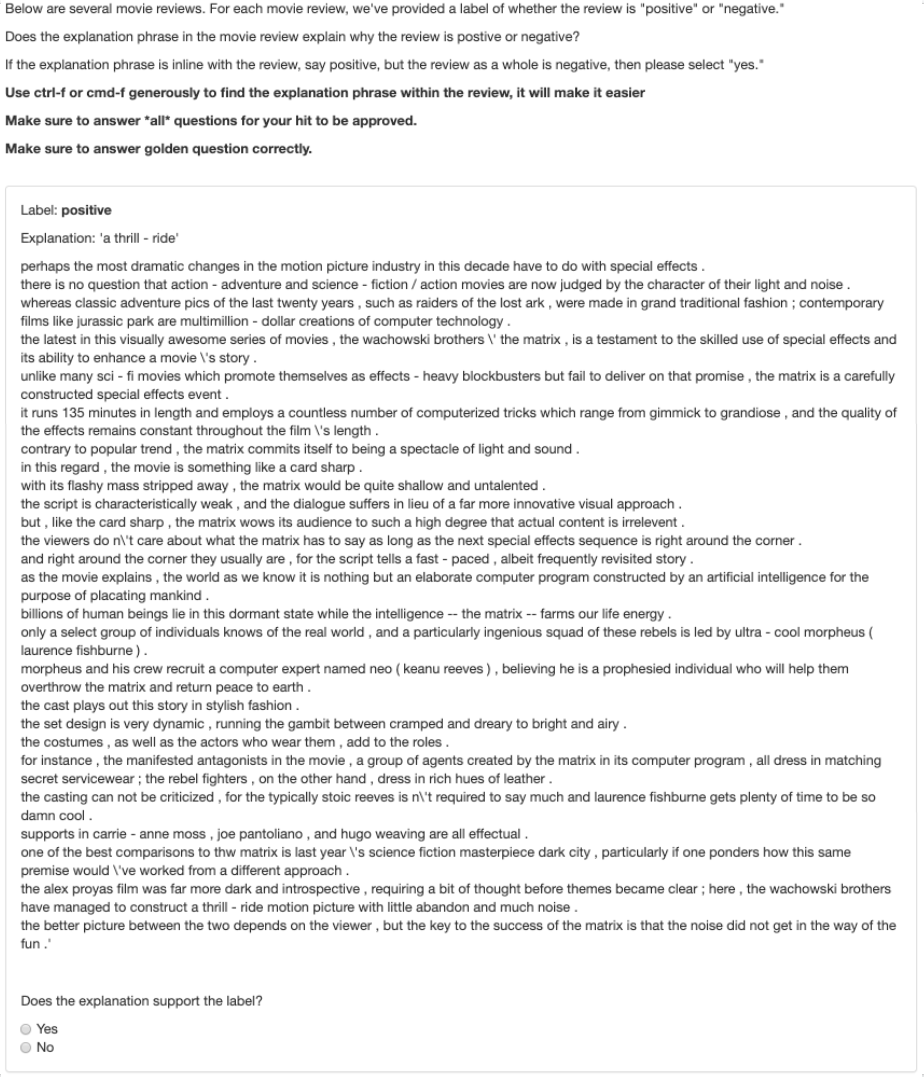}
    \caption{GUI for Movie Reviews (also IMDB). The explanation is from the validation set of the dataset.}
    \label{fig:gui-movie}
  \end{center}
  \vskip -0.2in
\end{figure*}

\begin{figure*}[ht]
  \begin{center}
    \includegraphics[width=0.8\textwidth]{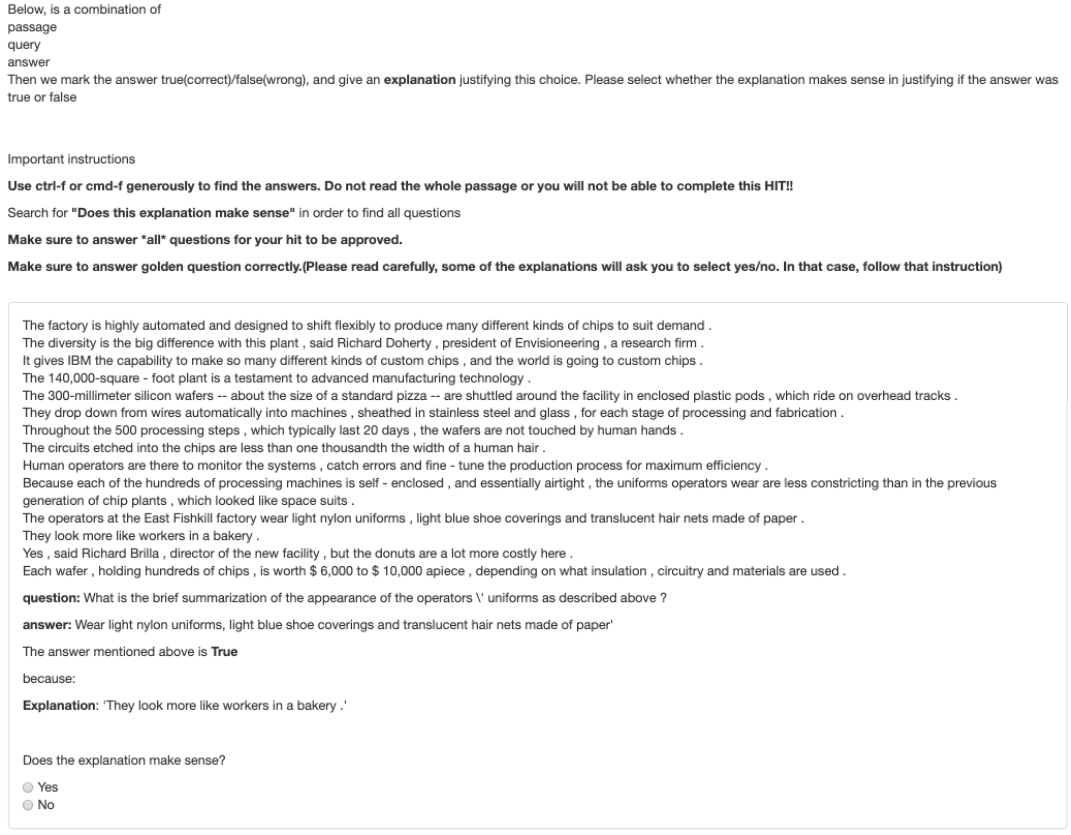}
    \caption{GUI for MultiRC. The explanation is from the validation set of the dataset.}
    \label{fig:gui-multirc}
  \end{center}
  \vskip -0.2in
\end{figure*}

Figures \ref{fig:gui-mnli} through \ref{fig:gui-multirc} show the GUIs we posted to MTurk to evaluate different datasets. When the dataset includes a human-generated explanation for the label, we perform human evaluation on both the provided explanation and generated explanation. Several datasets share the same MTurk setup as noted in the captions. 

\newpage

\end{document}